# Gait2Hip-60: A Unified Deep Learning Benchmark for Predicting Hip Muscle Forces and Joint Moments from Multi-Cadence Gait Kinematics

Jiaqi Zhang[1], Ji Hou[1], Qing Sun[1], Xianzhi Gao[2], Bo Huo[1,3*]

[1] Capital University of Physical Education and Sports.

[2] Beijing Institute of Technology.

[3]Beijing Key Laboratory of Interdisciplinary Intelligent Technologies of Sports, Medicine and Engineering.

zhangjiaqi2023@cupes.edu.cn

**Abstract**

Estimating hip muscle forces and joint moments during gait typically relies on musculoskeletal simulation, which is informative but time-consuming and difficult to apply in clinical settings. This study developed a deep learning framework to predict these hip dynamics parameters directly from lower-limb gait kinematics and compared three representative sequence models under a unified protocol. Gait data were collected from 60 healthy adults under three metronome-guided cadence conditions. Ten bilateral lower-limb joint angles were used as inputs, and OpenSim-derived hip muscle forces and hip joint moments were used as reference outputs. Three deep learning models of LSTM, Transformer, and Mamba were trained and evaluated using the same subject-level split, preprocessing pipeline, and metrics. The best model was then directly tested on an external cohort of 9 patients with osteonecrosis of the femoral head (ONFH) without retraining. In the healthy-subject benchmark, Transformer achieved the best subject-level mean performance for both hip muscle force prediction (RMSE = 1.33 N/kg, MAE = 0.57 N/kg, $R^2$ = 0.819) and hip joint moment prediction (RMSE = 0.11 Nm/kg, MAE = 0.07 Nm/kg, $R^2$ = 0.862), with similar advantages across walking cadences. In zero-shot external validation, Transformer retained moderate predictive ability in ONFH for hip muscle force prediction (RMSE = 1.51 N/kg, MAE = 0.70 N/kg, $R^2$ = 0.537) and hip joint moment prediction (RMSE = 0.17 Nm/kg, MAE = 0.12 Nm/kg, $R^2$ = 0.569). These findings support the feasibility of estimating hip dynamics from gait kinematics, identify Transformer as a strong baseline, and highlight the need for broader pathological validation and improved generalization before clinical application.

## 1. Introduction

The hip joint is a key weight-bearing joint during walking and plays an important role in pelvic stability, forward progression, and load transfer between the trunk and lower limbs [1,2]. Hip muscle forces and joint moments are therefore important internal biomechanical variables for understanding gait function, abnormal movement patterns, and compensatory strategies in both research and clinical settings [2,3]. However, direct measurement of these variables is difficult, and their estimation often depends on complex experimental procedures and time-consuming analyses [4,5].

Musculoskeletal modeling remains one of the main approaches for estimating muscle forces and joint moments [3,4,6]. Platforms such as OpenSim can generate biomechanically meaningful outputs from motion data, external forces, and musculoskeletal geometry through inverse kinematics, inverse dynamics, and optimization procedures [4,6]. These methods are widely used in gait analysis and also provide reference labels for supervised learning studies. However, the full workflow usually requires motion capture, force plates, subject-specific scaling, and multiple processing steps, which limits its use in rapid assessment, repeated monitoring, and routine clinical practice [3,5].

To improve efficiency and accessibility, recent studies have explored machine learning and deep learning methods for predicting lower-limb dynamics directly from signals that are easier to obtain, such as kinematics [7], inertial sensor data, and surface electromyography [8–10]. Among these tasks, joint moment prediction has received substantial attention, and a range of models have been reported, including recurrent networks, convolution-based models, and attention-based architectures [10]. By comparison, direct prediction of muscle forces remains less prevalent, even though muscle force profiles may better reflect neuromuscular control and load-sharing strategies during walking [11,12]. In addition, previous studies differ considerably in signal type, movement task, sample size, output definition, and evaluation strategy, making it difficult to compare model performance under a unified setting [8–12].

Another important challenge is generalization. Many existing studies are still limited to healthy participants or to in-distribution evaluation [8–10]. External validation in patient populations is relatively uncommon, although pathological gait often involves altered kinematics, abnormal load transfer, and compensatory strategies that may change the relationship between observable motion and internal dynamics [9,13]. Osteonecrosis of the femoral head, ONFH, is a clinically relevant example because it is associated with structural hip damage and clear gait abnormalities. It therefore provides a useful test case for evaluating whether a model trained on healthy gait data can retain predictive value under distribution shift [13,14].

To address these limitations, this study developed a deep learning framework to predict hip muscle forces and hip joint moments directly from lower-limb joint angle sequences. Three representative sequence models, LSTM, Transformer, and Mamba, were compared under the same training and evaluation protocol [15–17]. The study had three aims: first, to establish a multi-cadence gait dataset from 60 healthy subjects; second, to provide a unified benchmark for predicting both hip muscle forces and hip joint moments; third, to test the best model in patients with ONFH using strict zero-shot external validation for both prediction tasks. Ultimately, the study aimed to identify a strong baseline model for kinematics-to-dynamics prediction and to evaluate its transferability to a clinically different population (Fig. 1).

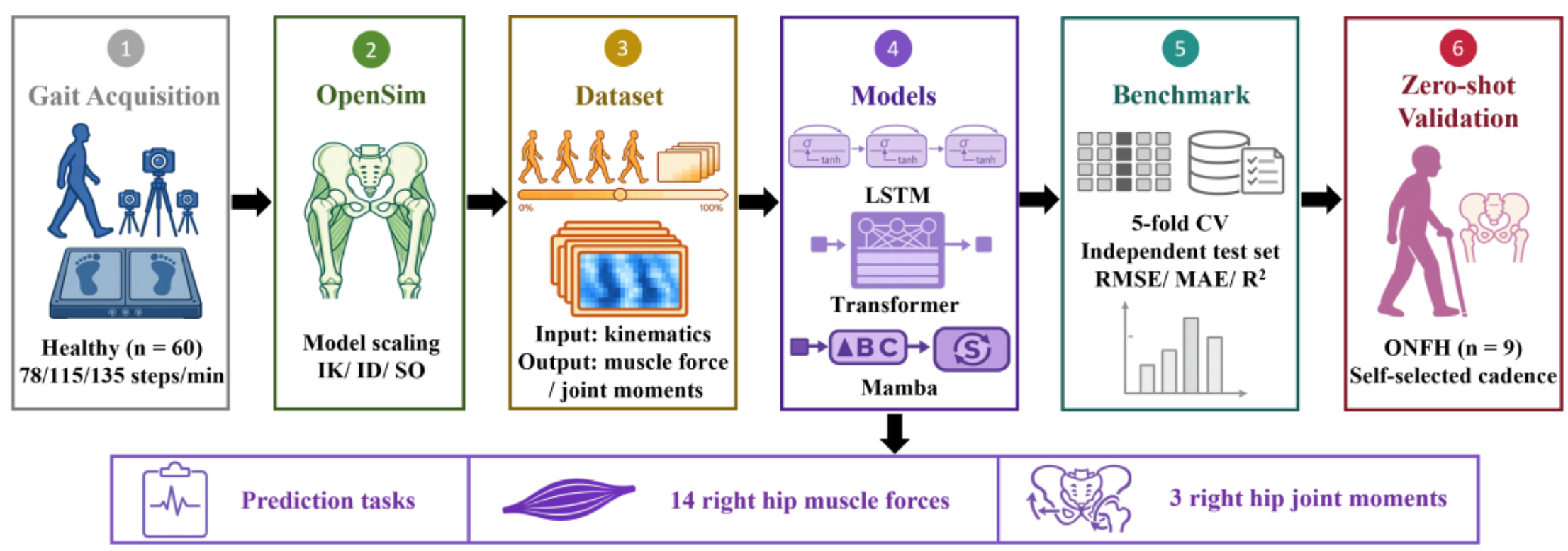

**Figure 1.** Overview of the study workflow for predicting hip muscle forces and joint moments from multi-cadence gait kinematics.

## 2. Materials and Methods

### 2.1 Participants and Data Acquisition

This study was approved by the Ethics Committee of Capital University of Physical Education and Sports (Approval No. 2024A120), and all participants provided written informed consent before data collection. Two cohorts were included: a healthy cohort and an ONFH cohort. Participant characteristics are summarized in Table 1. The healthy cohort included 60 young adults without lower-limb musculoskeletal disorders, neurological disease, or recent injury affecting gait. The ONFH cohort included 9 patients with clinically diagnosed stage II ONFH [14]. The healthy cohort was used for model training and testing, whereas the ONFH cohort was reserved only for external validation.

Gait data were collected on an indoor level walkway using 41 retroreflective markers, 8 infrared motion-capture cameras (Prime 13, OptiTrack, USA; 200 Hz), and 4 synchronized force plates (9260AA, Kistler, Switzerland; 1000 Hz). The marker set and experimental setup are shown in Fig. 2. In the healthy cohort, walking cadence was guided by a metronome at 78, 115, and 135 steps/min (one beat per step), creating three walking conditions intended to relatively slow medium and fast walking. Each participant completed six trials per condition, and four high-quality trials were retained after quality control, resulting in 713 valid trials. The ONFH cohort walked at a self-selected comfortable cadence.

**Table 1.** Participant characteristics of the healthy and ONFH cohorts. Height and mass are presented as mean ± SD, age is presented as range.

| Cohort | N | Sex (M/F) | Age (years) | Height (m) | Mass (kg) | Walking condition (steps/min) |
|---|---|---|---|---|---|---|
| Healthy | 60 | 30/30 | 18 - 30 | 1.71 ± 0.08 | 66.35 ± 12.61 | 78, 115, and 135 |
| ONFH | 9 | 6/3 | 38 - 68 | 1.71 ± 0.06 | 70.99 ± 14.73 | self-selected |

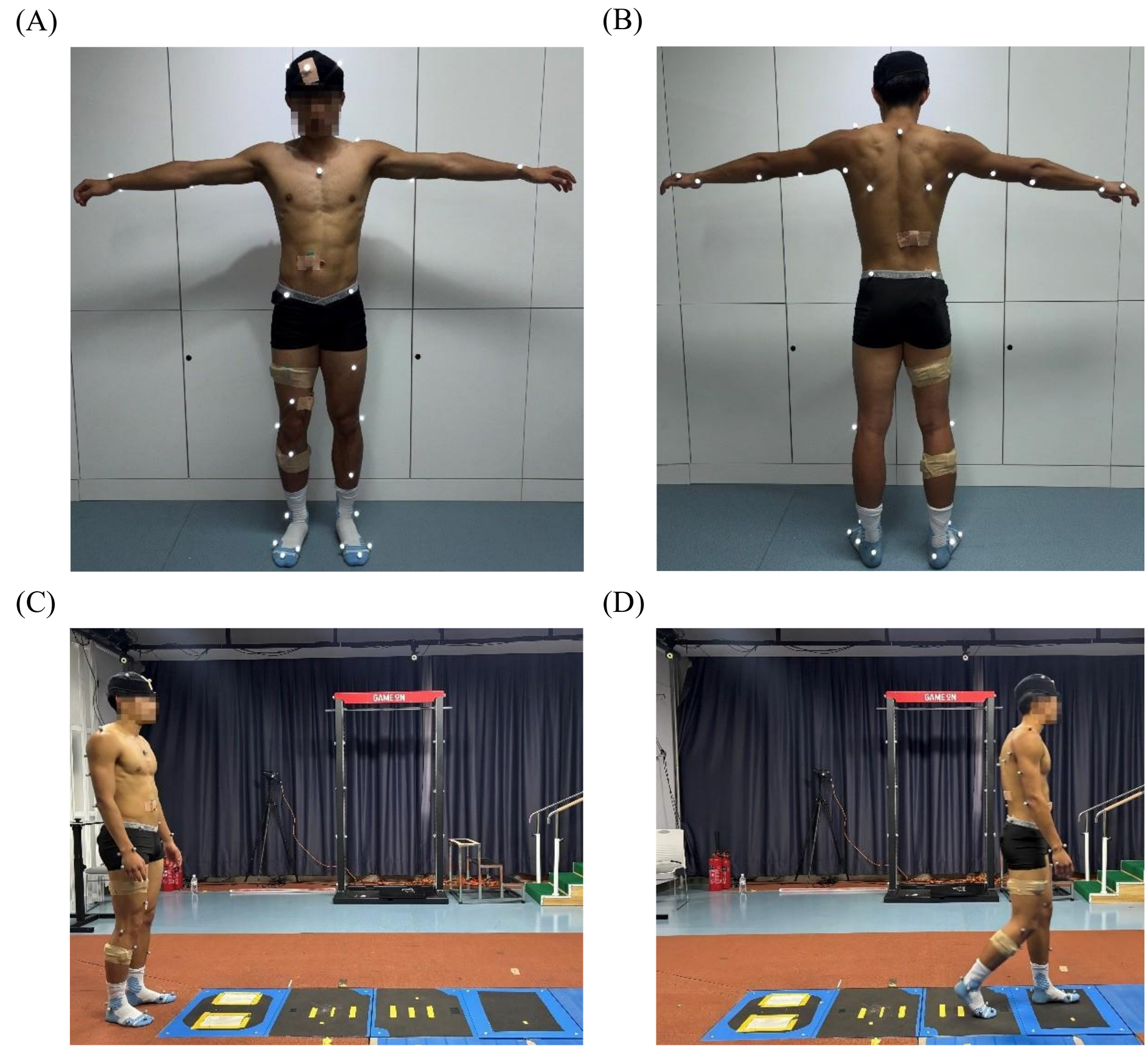

**Figure 2.** Marker placement and experimental setup for data acquisition. (A) Anterior view during static calibration. (B) Posterior view during static calibration. (C) Lateral view during gait acquisition. (D) Representative walking trial during gait acquisition.

### 2.2 Musculoskeletal Modeling and Dataset Preparation

Musculoskeletal modeling was performed within a unified OpenSim workflow [4], as shown in Fig. 3. The full-body musculoskeletal model developed by Rajagopal et al. [18] was adopted as the base model. Subject-specific scaling was carried out from the static standing trial using the Automatic Scaling Tool [19]. Scaling quality was assessed by marker fitting errors, with RMSE maintained below 0.004 m.

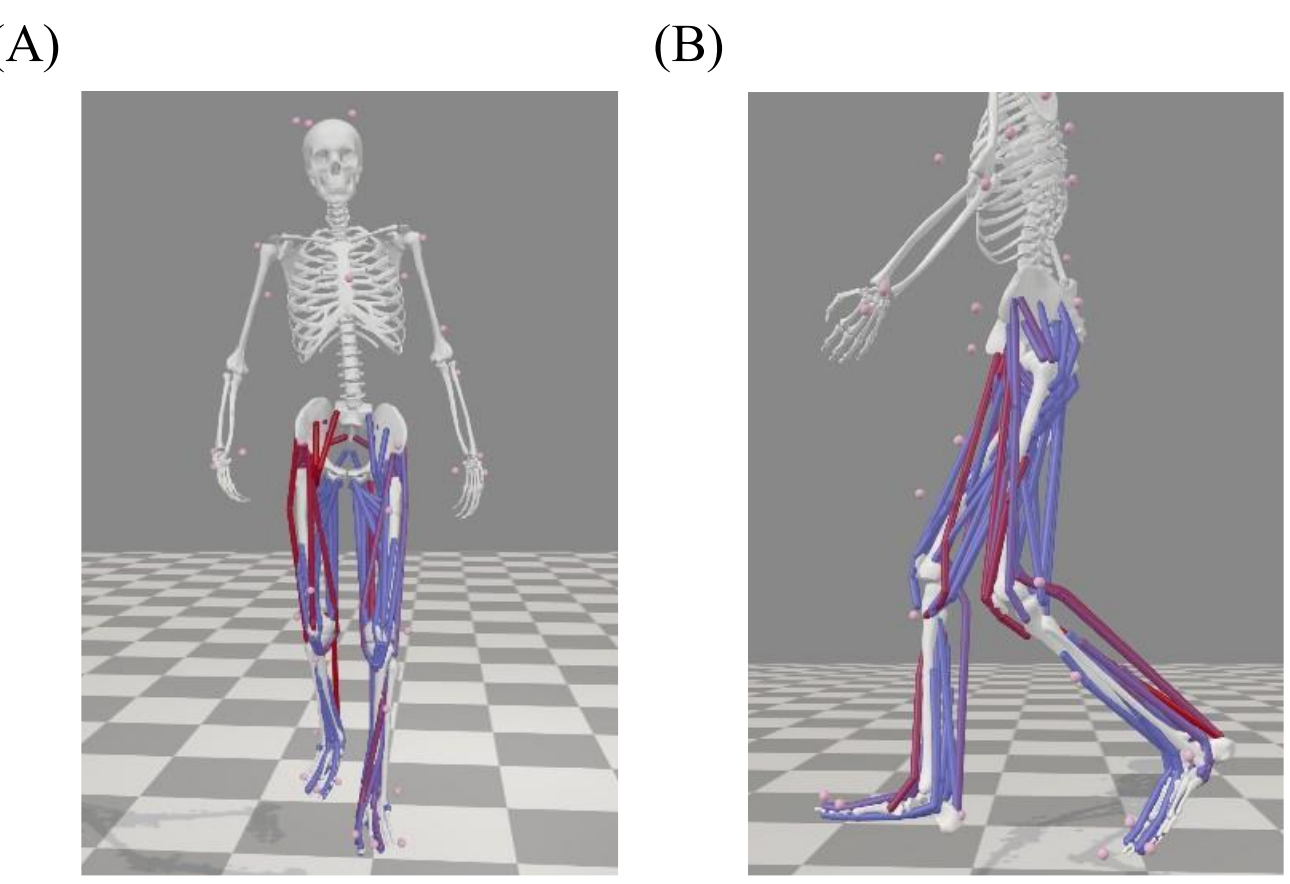

Figure 3. OpenSim musculoskeletal model used for gait analysis. (A) Frontal view. (B) Lateral view.

After scaling, inverse kinematics (IK) was performed frame by frame to estimate lower-limb joint angles. The IK problem was formulated as a weighted least-squares marker-tracking problem and the function , denoted as $J_{IK}(q)$, was defined as:

$$J_{IK}(q) = \sum_{i=1}^{N} w_i \left\| x_i^{exp} - x_i^{model}(q) \right\|^2 \quad (1)$$

where $q$ denotes the generalized coordinates, $N$ is the number of markers, $w_i$ is the weight assigned to the $i$-th marker, and $x_i^{exp}$ and $x_i^{model}(q)$ are the experimental and model marker positions, respectively.

Inverse dynamics (ID) was then performed together with ground reaction forces to compute joint moments. The ID problem was described by the multibody equations of motion which defined as:

$$M(q)\ddot{q} + C(q, \dot{q}) + G(q) = \tau + F_{ext} \quad (2)$$

where $M(q)$ is the mass matrix, $\ddot{q}$ is the generalized acceleration vector, $C(q, \dot{q})$ represents the Coriolis and centrifugal terms, $G(q)$ is the gravity term, $\tau$ denotes the joint moments, and $F_{ext}$ represents the external forces.

Hip muscle forces were subsequently estimated using the static optimization (SO) tool implemented in OpenSim [20], which minimizes the sum of squared muscle activations while satisfying joint moment constraints. The SO objective function, denoted as $J_{SO}(a)$, was defined as:

$$J_{SO}(a) = \sum_{m=1}^{M} a_m^2 \quad (3)$$

subject to:

$$R(q)F_m = \tau, \qquad 0 \le a_m \le 1 \quad (4)$$

where $a$ is the vector of muscle activation, $a_m$ is the activation of the $m$-th muscle, $M$ is the total number of muscles, $F_m$ denotes muscle force, and $R(q)$ is the moment-arm matrix.

For dataset construction, each sample corresponded to one gait cycle, defined from one ipsilateral initial contact to the next ipsilateral initial contact [21]. All signals were normalized to 0-100% of the gait cycle and resampled to 180 frames using linear interpolation. Ten bilateral lower-limb kinematic variables were used as inputs, including bilateral hip, knee, and ankle joint angles. Hip muscle forces and joint moments were used as outputs.

To improve training stability, the input was standardized using the training-set statistics:

$$X_{std} = \frac{X - \mu_{train}}{\sigma_{train}} \quad (5)$$

The $\mu_{train}$ and $\sigma_{train}$ were applied to the validation, test, and ONFH datasets to avoid information leakage. Both muscle forces and joint moments were normalized by body mass:

$$F_{norm} = \frac{F}{m} \quad (6)$$

$$M_{norm} = \frac{M}{m} \quad (7)$$

where $m$ is body mass.

**2.3 Model Training and Evaluation**

Three representative sequence models were compared: LSTM [15], Transformer [17], and Mamba [16]. All models used the same kinematic inputs and were trained under the same protocol. Two related prediction tasks were considered: hip muscle force prediction and hip joint moment prediction.

Data splitting was performed at the subject level. In the healthy cohort, 80% of participants were assigned to the training pool and the remaining 20% were reserved as an independent test set. Within the training pool, model evaluation was conducted using 5-fold cross-validation at the subject level.

Training settings were kept identical across models. Mean squared error (MSE) was used as the loss function:

$$MSE(y_i, \hat{y}_i) = \frac{1}{N}\sum_{i=1}^{N}(y_i - y_i)^2 \tag{8}$$

where $y_i$ and $\hat{y}_i$ denote the reference and predicted values, respectively, and $N$ is the number of evaluated points.

Model parameters were optimized using AdamW [22], with a learning rate of $1 \times 10^{-4}$, weight decay of $1 \times 10^{-5}$, a batch size of 64, and a maximum of 300 epochs. Gradient clipping was applied with a maximum norm of 0.5. Early stopping was applied during cross validation, and the final model was retrained using the median best epoch across folds. Random seeds were fixed throughout all experiments to improve reproducibility. All models were implemented in PyTorch and trained on a single NVIDIA GeForce RTX 2080 Ti GPU.

Model performance was evaluated using RMSE, mean absolute error (MAE), and the coefficient of determination ($R^2$):

$$RMSE = \sqrt{\frac{1}{N}\sum_{i=1}^{N}(y_i - \hat{y}_i)^2} \tag{9}$$

$$MAE = \frac{1}{N}\sum_{i=1}^{N}|y_i - \hat{y}_i| \tag{10}$$

$$R^2 = 1 - \frac{\sum_{i=1}^{N}(y_i - \hat{y}_i)^2}{\sum_{i=1}^{N}(y_i - \bar{y})^2} \tag{11}$$

where $\bar{y}$ is the mean of the reference values. Pairwise model comparisons were performed using Wilcoxon signed-rank tests with Holm correction for multiple comparisons.

### 2.4 External Validation in ONFH

After benchmarking on the healthy cohort, the best model was directly applied to the ONFH cohort for external validation of muscle force and joint moment prediction. ONFH data were excluded from model training, resulting in a strict zero-shot validation setting. The ONFH data were processed using the same musculoskeletal modeling and preprocessing pipeline as the healthy data, including gait cycle segmentation, resampling, input standardization, and body mass normalization. The only difference was that the ONFH participants walked at a self-selected comfortable cadence.

## 3. Results

### 3.1 Hip Muscle Force Prediction

All three sequence models were able to predict hip muscle forces from gait kinematics in the healthy-subject test set. Among them, the Transformer achieved the best subject-level mean performance, with an RMSE of 1.33 N/kg, an MAE of 0.57 N/kg, and an $R^2$ of 0.819. In comparison, the LSTM yielded an RMSE of 1.51 N/kg, an MAE of 0.66 N/kg, and an $R^2$ of 0.775, whereas Mamba yielded an RMSE of 1.39 N/kg, an MAE of 0.60 N/kg, and an $R^2$ of 0.804. Pairwise comparisons showed that the Transformer significantly outperformed the LSTM across all three overall metrics, whereas its advantage over Mamba reached significance for MAE only.

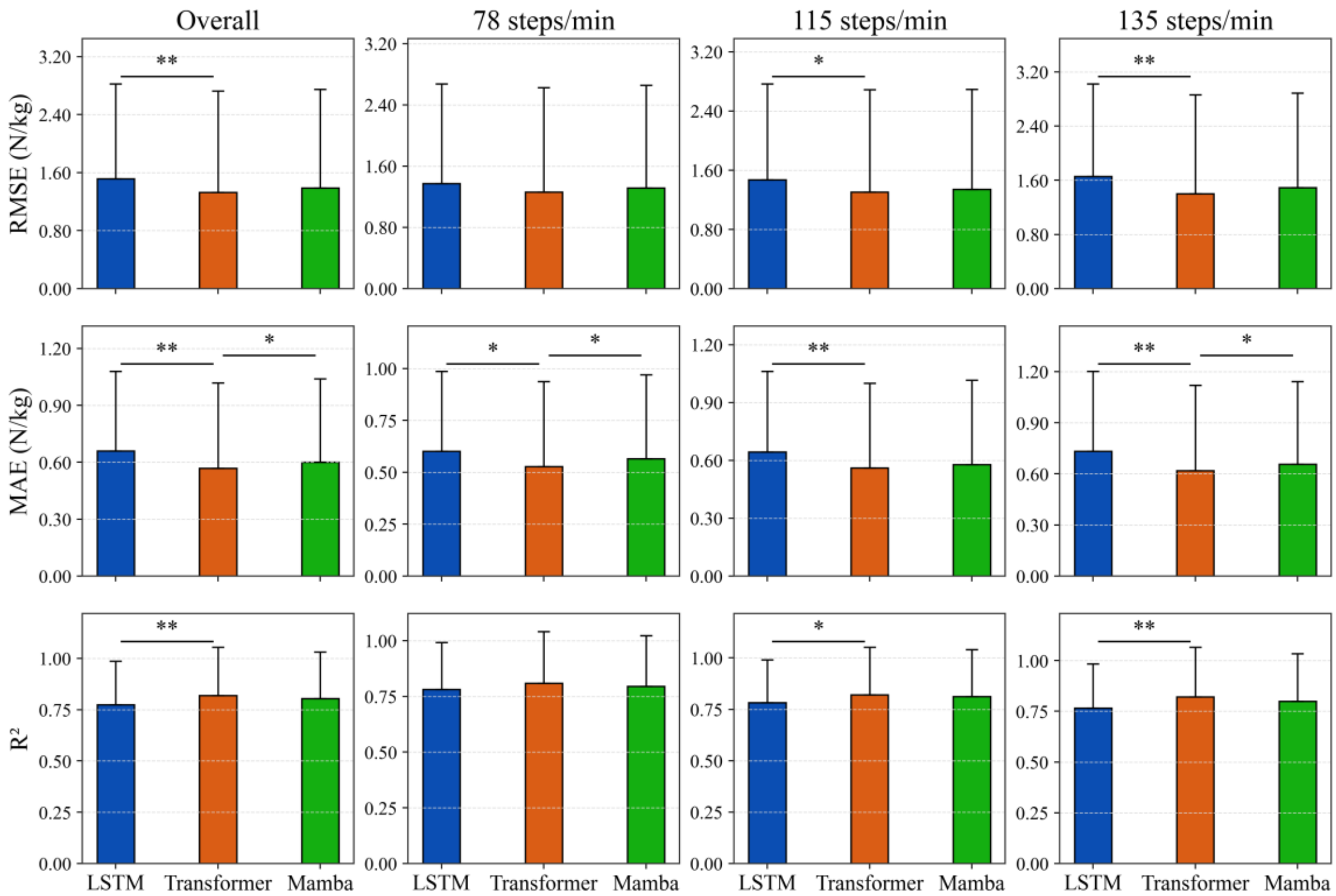

Figure 4. Comparison of LSTM, Transformer, and Mamba for hip muscle force prediction in the healthy-subject test set under overall and three cadence conditions. Horizontal brackets indicate significant pairwise differences after Holm correction (* $p < 0.05$, ** $p < 0.01$).

This pattern was generally maintained across three cadence conditions. As shown in Fig. 4, the Transformer remained the best-performing model across the three cadence conditions (78, 115, and 135 steps/min), with RMSE values of 1.26, 1.31, and 1.40 N/kg, MAE values of 0.53, 0.56, and 0.62 N/kg, and $R^2$ values of 0.808, 0.820, and 0.822, respectively. In the 78 steps/min condition, significant differences involving the Transformer were observed only for MAE. In the 115 steps/min condition, the Transformer significantly outperformed the LSTM across all three metrics, whereas the differences between Transformer and Mamba were not statistically significant. In the 135 steps/min condition, the Transformer again significantly outperformed the LSTM across all three metrics, while its advantage over Mamba remained significant only for MAE.

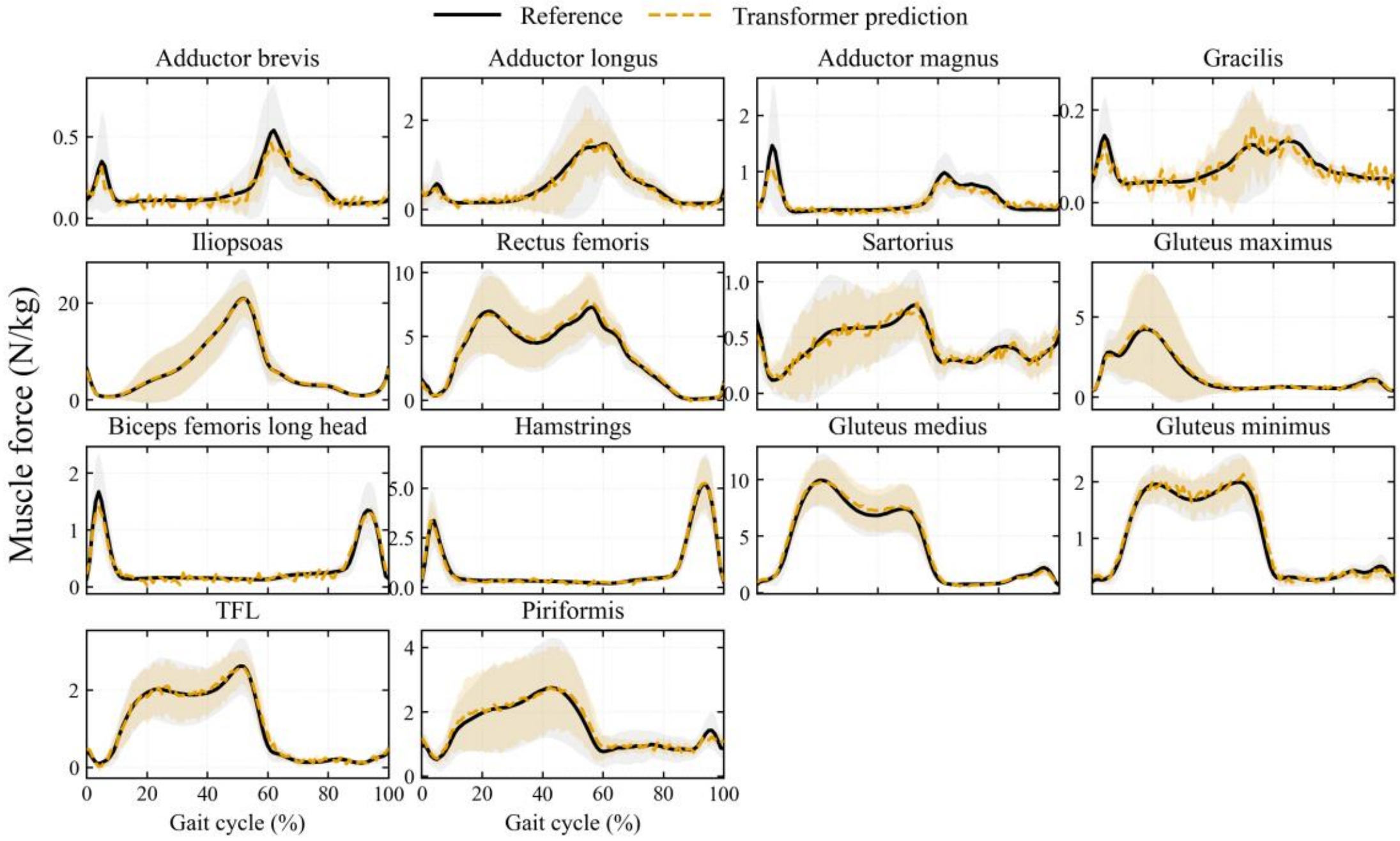

Figure 5. Waveform comparison of 14 right hip muscle forces in the healthy-subject test set. Black solid lines show the reference values, yellow dashed lines show the predictions, and shaded areas indicate mean ± standard deviation.

As shown in Fig. 5, the Transformer, which showed the best overall performance, captured the overall waveform shapes and major peaks of most hip muscle forces across the gait cycle. The predicted curves closely matched the reference profiles across the gait cycle and preserved the main temporal patterns. Differences were mainly seen near the peaks and during phase transitions, and were more obvious for muscles with lower force levels or larger differences between subjects. These qualitative findings were consistent with the superior quantitative performance of the Transformer under the unified benchmark setting.

**3.2 Hip Joint Moment Prediction**

A similar trend was observed for hip joint moment prediction. The Transformer again achieved the best subject-level mean performance in the healthy-subject test set, with an RMSE of 0.11 Nm/kg, an MAE of 0.07 Nm/kg, and an $R^2$ of 0.862. The corresponding values were 0.12 Nm/kg, 0.08 Nm/kg, and 0.849 for the LSTM, and 0.13 Nm/kg, 0.08 Nm/kg, and 0.815 for Mamba. Pairwise comparisons showed that the Transformer significantly outperformed Mamba across all three overall metrics, whereas the differences between Transformer and LSTM were not statistically significant.

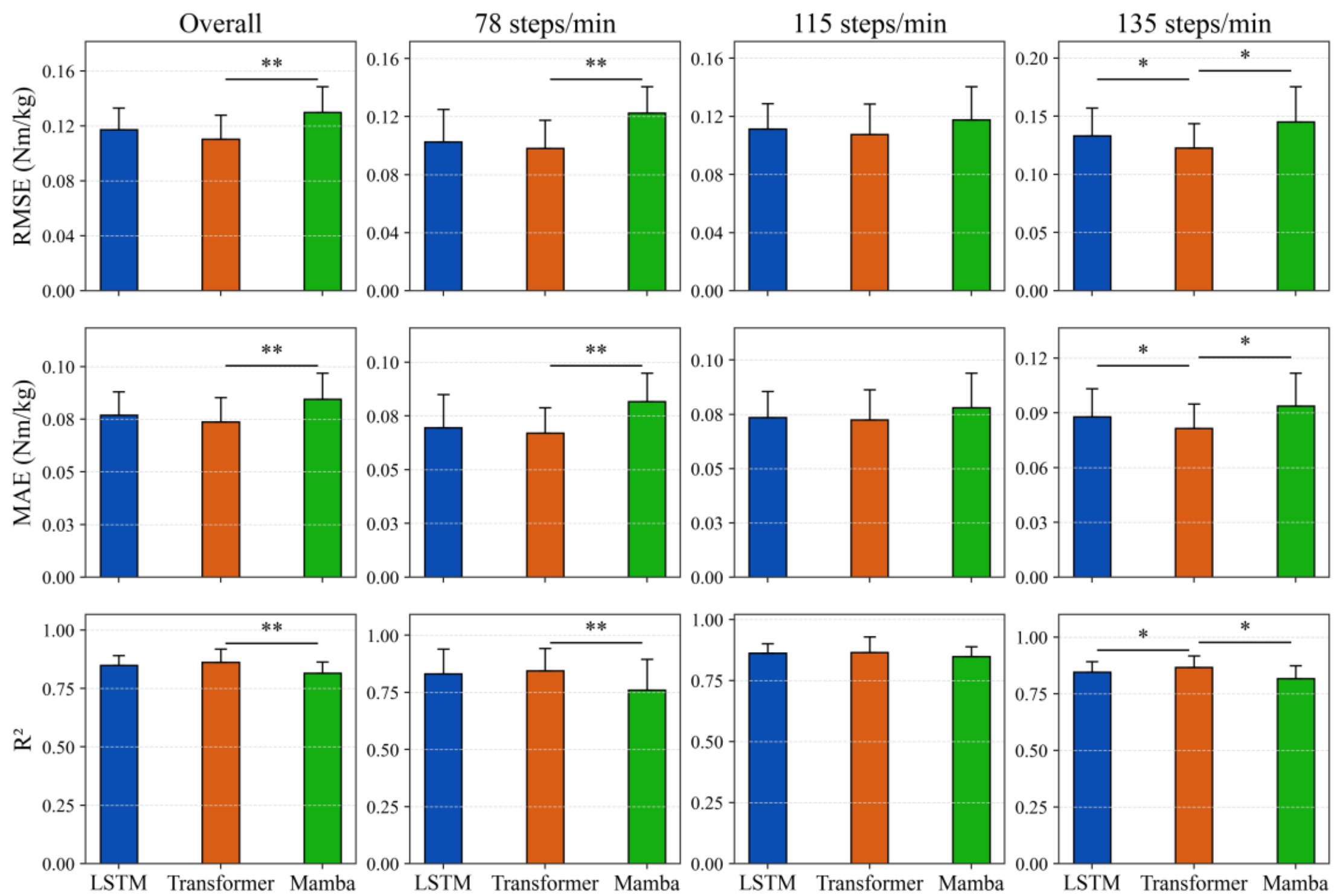

Figure 6. Comparison of LSTM, Transformer, and Mamba for hip joint moment prediction in the healthy-subject test set under overall and three cadence conditions. Horizontal brackets indicate significant pairwise differences after Holm correction (* $p < 0.05$, ** $p < 0.01$).

The same general ranking was observed across walking cadences, as shown in Fig. 6. Across the three cadence conditions, the Transformer remained the best-performing model, with RMSE values of 0.10, 0.11, and 0.12 Nm/kg, MAE values of 0.07, 0.07, and 0.08 Nm/kg, and $R^2$ values of 0.844, 0.865, and 0.867, respectively. In the 78 steps/min condition, the Transformer significantly outperformed Mamba, whereas no significant differences were found between the Transformer and LSTM. In the 115 steps/min condition, none of the pairwise comparisons reached statistical significance. In the 135 steps/min condition, the Transformer significantly outperformed both LSTM and Mamba across all three metrics. These results suggest that the relative advantage of the Transformer was most pronounced at 135 steps/min, remained evident against Mamba at 78 steps/min walking, and became less distinct at 115 steps/min.

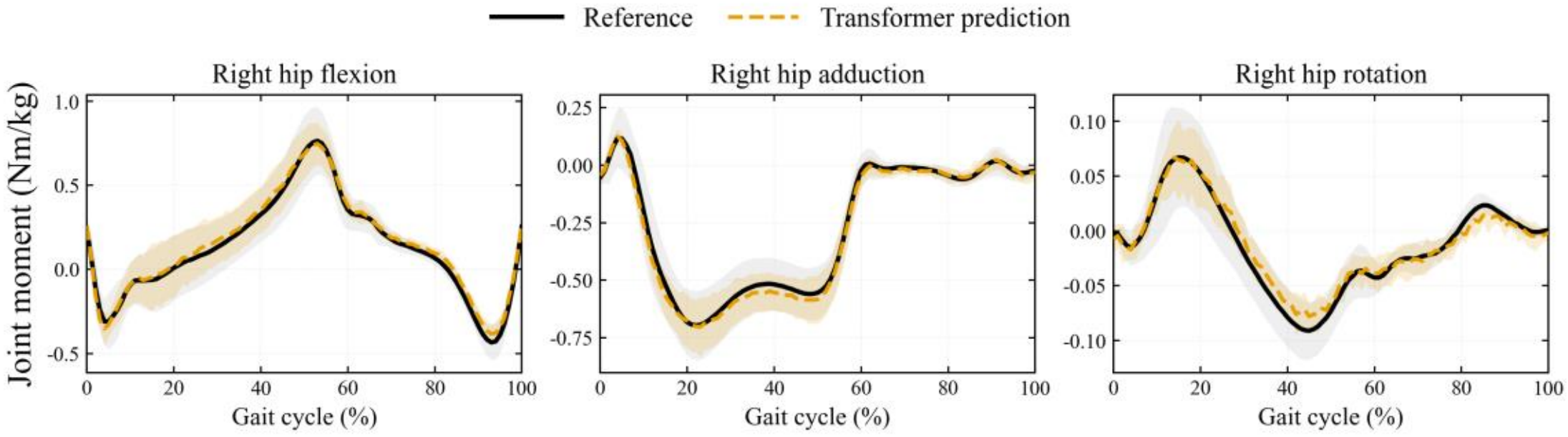

Figure 7. Waveform comparison of 3 right hip joint moments in the healthy-subject test set. Black solid lines show the reference values, yellow dashed lines show the predictions, and shaded areas indicate mean ± standard deviation.

Waveform comparisons of the best model further supported the quantitative results. As shown in Fig. 7, the Transformer captured the main temporal patterns of all three hip joint moment components across the gait cycle. The predicted curves were generally closer for hip flexion and hip adduction moments, with the major peaks and overall waveform shapes well preserved. In contrast, the hip rotation moment showed relatively larger local deviations, especially in regions with smaller magnitudes and more variable waveform patterns. In summary, these findings were consistent with the superior performance of the Transformer in hip joint moment prediction.

### 3.3 External Validation in ONFH

To further assess external generalization, the best-performing model identified in the healthy-subject benchmark, namely the Transformer, was directly applied to the ONFH cohort under a strict zero-shot setting. For hip muscle force prediction, the Transformer achieved a subject-level mean RMSE of 1.51 N/kg, an MAE of 0.70 N/kg, and an $R^2$ of 0.537. For hip joint moment prediction, the corresponding values were 0.17 Nm/kg, 0.12 Nm/kg, and 0.569. These results are summarized in Table 2 and 3.

**Table 2.** Comparison of Transformer performance between the healthy cohort and the ONFH cohort in hip muscle force prediction. Values are presented as mean ± SD.

| Cohort | RMSE (N/kg) | MAE (N/kg) | $R^2$ |
|---|---|---|---|
| Healthy | 1.33 ± 1.40 | 0.57 ± 0.45 | 0.819 ± 0.235 |
| ONFH | 1.51 ± 0.44 | 0.70 ± 0.19 | 0.537 ± 0.355 |

**Table 3.** Comparison of Transformer performance between the healthy cohort and the ONFH cohort in hip joint moment prediction. Values are presented as mean ± SD.

| Cohort | RMSE (Nm/kg) | MAE (Nm/kg) | $R^2$ |
|---|---|---|---|
| Healthy | 0.11 ± 0.02 | 0.07 ± 0.01 | 0.862 ± 0.057 |
| ONFH | 0.17 ± 0.05 | 0.12 ± 0.04 | 0.569 ± 0.198 |

Performance distributions are illustrated in Fig. 8, where the central line represents the median and the box indicates the interquartile range. Waveform comparisons further showed that the healthy-trained Transformer still captured the main phase-dependent patterns of several hip muscle forces and the overall temporal trends of the three hip joint moment components (Figs. 9).

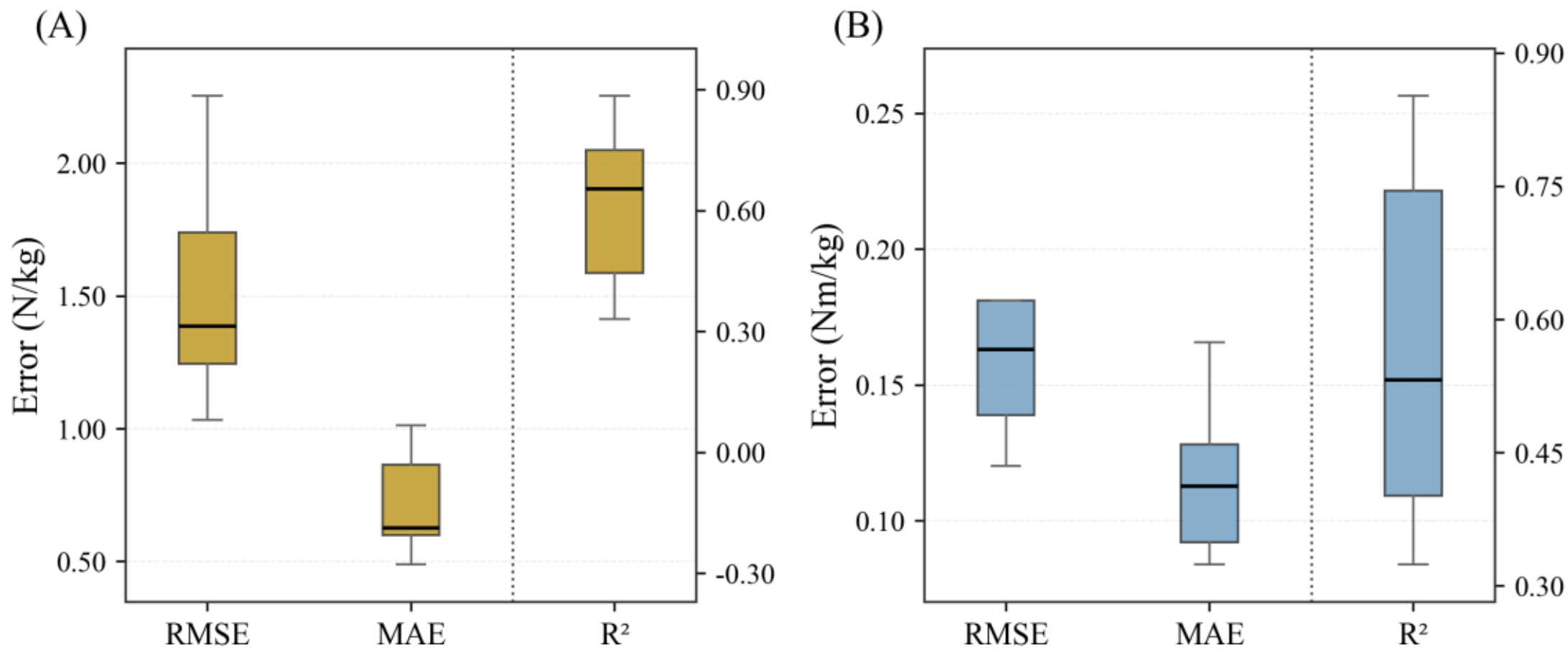

Figure 8. Performance of the Transformer in the ONFH cohort. Boxes indicate interquartile ranges and center lines indicate medians. (A) Hip muscle force prediction. (B) Hip joint moment prediction.

Overall waveform patterns were generally preserved throughout most of the gait cycle, although larger deviations were observed near peak values and transition periods. Compared with the healthy cohort, performance declined in the ONFH cohort for both hip muscle force and joint moment prediction, suggesting reduced generalization under pathological gait conditions. Nevertheless, the waveform comparisons indicates that the Transformer retained moderate zero-shot predictive ability in the ONFH cohort.

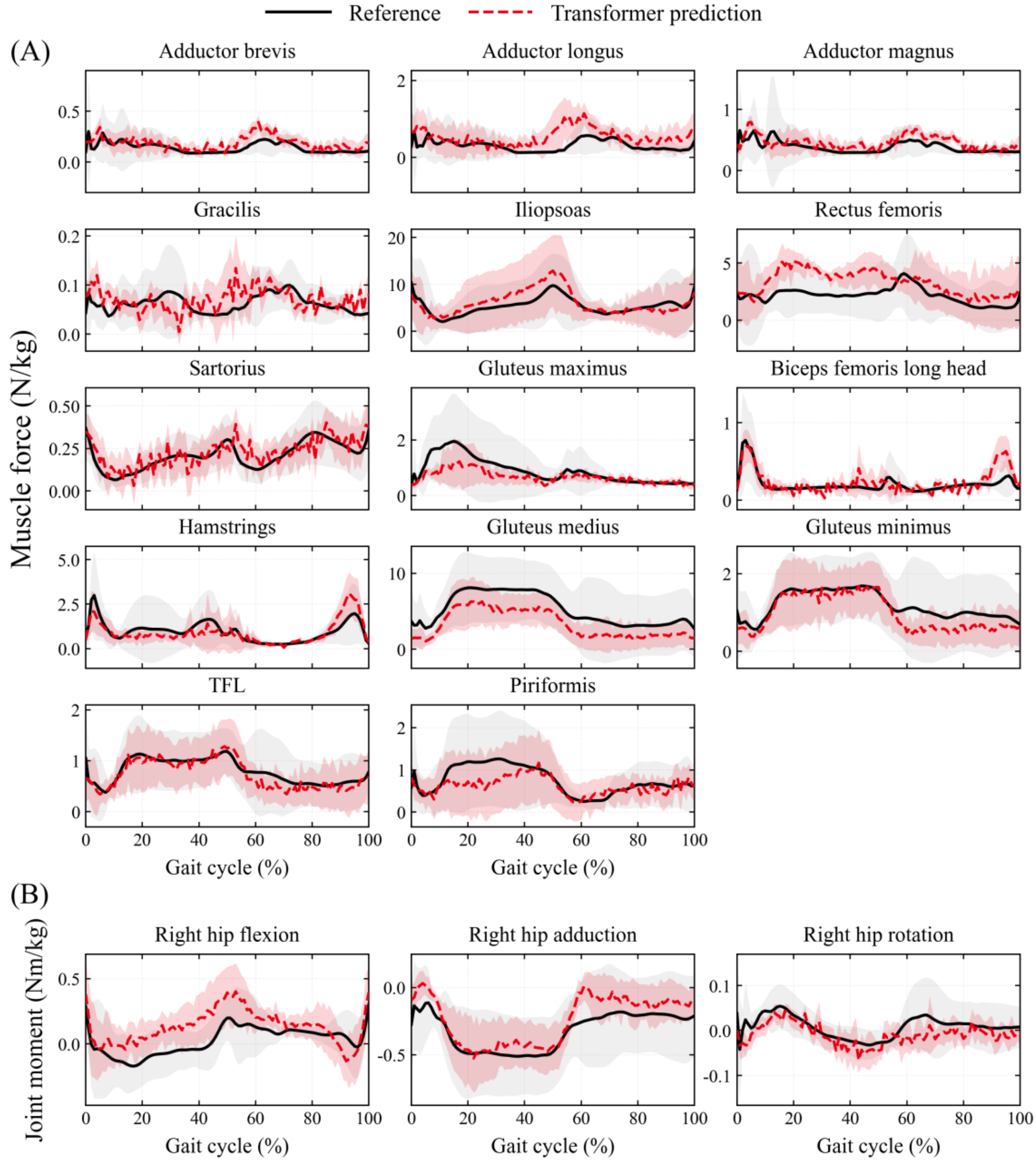

Figure 9. Waveform comparison of reference values and Transformer-predicted hip muscle forces (A) and hip joint moments (B) in the ONFH cohort. Shaded areas indicate mean ± standard deviation.

**4. Discussion**

This study developed a deep learning framework for estimating hip muscle forces and hip joint moments from gait kinematics and evaluated it using a unified benchmark in healthy subjects, followed by zero-shot external validation in patients with ONFH. The main results were consistent across tasks. Although all three models were able to learn the mapping from lower-limb kinematics to hip-related dynamics, the Transformer achieved the best subject-level mean performance for both muscle force and joint moment prediction. When directly applied to ONFH data, the healthy-trained model showed reduced accuracy but retained moderate predictive ability. Together, these findings support the feasibility of estimating clinically relevant hip dynamics from gait kinematics [10,11] while also highlighting the challenge of generalizing to pathological gait [13,23].

A key finding was the consistent superiority of the Transformer over LSTM and Mamba in the healthy-subject benchmark. This advantage was reflected not only in lower errors but also in better fitting with the reference waveforms and major patterns. One possible explanation is that hip muscle forces and joint moments depend on temporal relationships that extend across multiple gait phases rather than only on local frame-to-frame transitions. Compared with recurrent models, the Transformer can capture such long-range dependencies more directly through self-attention, which may be beneficial for modeling coordination between stance and swing in hip dynamics [24]. This may be especially important for muscle force prediction, because muscle force trajectories are high-dimensional and reflect both mechanical demand and neuromuscular control [11,25,26]. The same model ranking in the joint moment task further suggests that this advantage was not limited to a single output type.

Another important result was that the same framework remained effective when the target was changed from hip muscle forces to hip joint moments. This suggests that the model learned a relatively stable relationship between gait kinematics and hip dynamics rather than fitting only one specific output. This is meaningful because joint moments and muscle forces represent different but closely related levels of musculoskeletal function. Joint moments describe mechanical demand at the joint, whereas muscle forces provide a more detailed view of how that demand may be produced and shared across muscles. At the same time, the cadence-related differences observed in both tasks show that walking cadence remains an important source of variation and should be considered in future model training and evaluation [27,28].

The zero-shot external validation in ONFH further highlights both the value and the limitations of the proposed framework. Performance decline was expected because ONFH patients differ from healthy young subjects in age, function, compensation strategy, joint loading pattern, and movement variability, all of which may alter the relationship between observable kinematics and internal dynamics [13,23]. Even so, the Transformer retained moderate predictive ability for both hip muscle force and hip joint moment prediction, suggesting that it captured part of the shared biomechanical structure of walking rather than only healthy patterns. Several limitations should also be noted. First, the reference outputs were derived from an OpenSim-based workflow rather than direct in vivo measurements [4,6], and uncertainty in scaling, inverse dynamics, and static optimization may affect the training targets [5,29]. Second, the ONFH sample size was limited. In addition, the reference muscle forces in this cohort were derived from a general musculoskeletal model, which may be less reliable, whereas joint moments are less dependent on such assumptions [29,5]. Third, only kinematic signals were used as inputs; adding EMG [30], wearable sensor data [31,32], or force-related parameters may improve robustness [33]. Finally, the external validation was strictly zero-shot and did not include subject-specific calibration or domain adaptation. Future studies should therefore include larger pathological cohorts [34], broader external testing [35], and adaptation strategies to improve generalization while preserving biomechanical interpretability [36]. From an application perspective, the present framework should not be viewed as a replacement for full laboratory-based gait analysis in all settings. Rather, its main value is that it provides a faster kinematics-to-dynamics estimation strategy once gait kinematics are available, without requiring a full subject-specific musculoskeletal simulation workflow for every new analysis.

## 5. Conclusion

This study established multi-cadence dataset and a deep learning framework for predicting hip muscle forces and hip joint moments from gait kinematics and benchmarked three representative sequence models under a unified evaluation protocol. The Transformer achieved the best subject-level mean performance in both tasks, indicating that attention-based sequence modeling is well suited to hip dynamics estimation during walking. In addition, zero-shot external validation in ONFH showed that a model trained on healthy gait can retain partial predictive ability in a pathological population, although clear performance degradation remains under distribution shift. These findings support Transformer as a strong baseline for future kinematics-to-dynamics studies and highlight the need for broader pathological validation and improved generalization before clinical application.

## 6. Data and Code Availability

The processed gait dataset used in this study is publicly available in Zenodo: https://doi.org/10.5281/zenodo.20175768.

The code for model training and evaluation is publicly available at GitHub: https://github.com/ygsiete7/Gait2Hip-60.